\documentclass[sigconf]{acmart}

\usepackage{multirow}
\usepackage{subfig}
\usepackage{hyperref}

\AtBeginDocument{%
  \providecommand\BibTeX{{%
    \normalfont B\kern-0.5em{\scshape i\kern-0.25em b}\kern-0.8em\TeX}}}

\copyrightyear{2020} 
\acmYear{2020} 
\setcopyright{acmcopyright}\acmConference[MM '20]{Proceedings of the 28th ACM International Conference on Multimedia}{October 12--16, 2020}{Seattle, WA, USA}
\acmBooktitle{Proceedings of the 28th ACM International Conference on Multimedia (MM '20), October 12--16, 2020, Seattle, WA, USA}
\acmPrice{15.00}
\acmDOI{10.1145/3394171.3413984}
\acmISBN{978-1-4503-7988-5/20/10}

\begin{document}
\fancyhead{}

\title{Vision Meets Wireless Positioning: Effective Person Re-identification with Recurrent Context Propagation}


\author{Yiheng Liu, Wengang Zhou, Mao Xi, Sanjing Shen, Houqiang Li}
\affiliation{%
  \institution{CAS Key Laboratory of Technology in GIPAS, EEIS Department, \\ University of Science and Technology of China}}
\email{{lyh156, ximao, ssjxjx}@mail.ustc.edu.cn, {zhwg, lihq}@ustc.edu.cn}

\renewcommand{\shortauthors}{Liu and Tobin, et al.}

\begin{abstract}
  Existing person re-identification methods rely on the visual sensor to capture the pedestrians. The image or video data from visual sensor inevitably suffers the occlusion and dramatic variations of pedestrian postures, which degrades the re-identification performance and further limits its application to the open environment. On the other hand, for most people, one of the most important carry-on items is the mobile phone, which can be sensed by WiFi and cellular networks in the form of a wireless positioning signal. Such signal is robust to the pedestrian occlusion and visual appearance change, but suffers some positioning error. In this work, we approach person re-identification with the sensing data from both vision and wireless positioning. To take advantage of such cross-modality cues, we propose a novel recurrent context propagation module that enables information to propagate between visual data and wireless positioning data and finally improves the matching accuracy. To evaluate our approach, we contribute a new Wireless Positioning Person Re-identification (WP-ReID) dataset. Extensive experiments are conducted and demonstrate the effectiveness of the proposed algorithm. Code will be released at \href{https://github.com/yolomax/WP-ReID}{https://github.com/yolomax/WP-ReID}.
\end{abstract}




\maketitle

\section{Introduction}

Person re-identification aims to find a pedestrian of interest that may appear in different locations. Many research efforts dedicated to this task assume that the pedestrian is sensed by video cameras. 
With potential significant applications in large-scale video surveillance networks, this task has drawn increasing attention from both academia and industry in recent years. Impressive progress has been witnessed for image-based person re-identification \cite{varior2016gated, li2018harmonious, liu2016multi, chen2019abd, yang2018local, xie2020progressive} and video-based person re-identification \cite{mclaughlin2016recurrent, li2019multi, liu2019spatial, gu2019temporal}. Compared with the image-based methods, the extra motion information in the video sequences and the additional information between the frames enable the video-based methods to extract more discriminative and robust features, which makes video-based approaches more promising to handle the challenges caused by occlusion and blur in person re-identification.

\begin{figure}[t]
  \centering
  \includegraphics[width=0.95\linewidth]{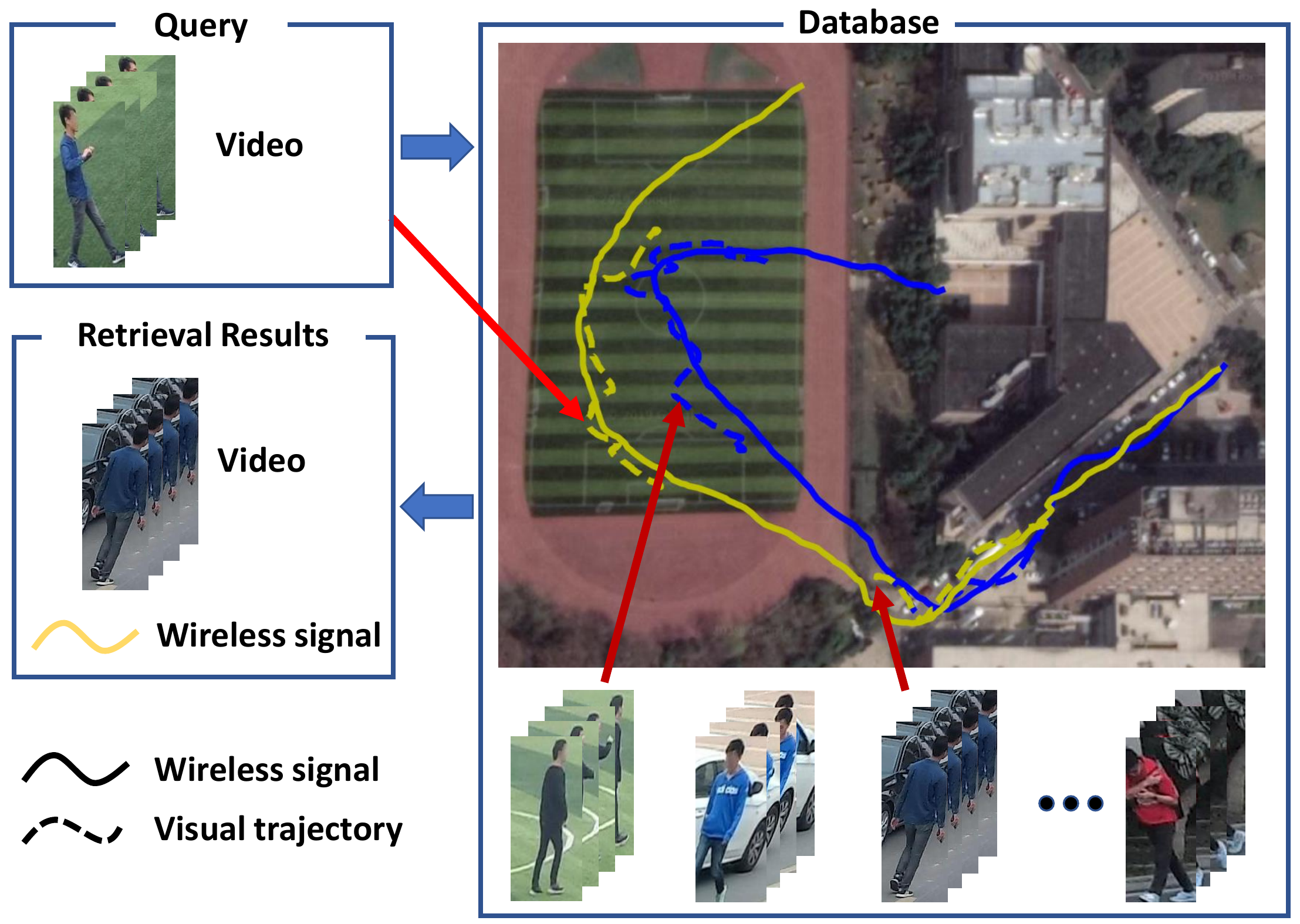}
  \caption{An overview of the problem studied in this work. Given a query video sequence, we not only seek to retrieve other videos with the same identity but also identify the corresponding wireless signal, \emph{i.e.,} the wireless positioning trajectory of the corresponding mobile phone taken by the same person. The solid line is the pedestrian trajectory obtained through wireless positioning. The dotted line is the trajectory of the pedestrian captured by cameras, which is obtained through coordinate mapping. Different colors represent different identities.}
  \label{fig:whole}
\end{figure}


Despite the remarkable achievements of the existing vision-based person re-identification methods, there are still some non-trivial issues. Current methods rely on the visual clue to discriminate different pedestrians. Although the visual clue provides a fine-grained description of pedestrians, it is very sensitive to occlusion, blur, and dramatic variations in pedestrian postures and clothes. Those interference factors can easily mislead the identification of pedestrians. On the other hand, mobile phones have become an indispensable item for most people to carry around. With WiFi or cellular networks, mobile phones can be easily sensed and positioned. Unlike vision data, such positioning signal is very robust to occlusion and appearance change, but is somewhat coarse-grained in describing a pedestrian. Witnessing the advantage and limitation of data from the vision sensor and wireless sensor, it is reasonable to fuse such heterogeneous cross-modality data to achieve more accurate and reliable person re-identification. 

Based on the above motivation, we propose to combine person re-identification with wireless positioning to further improve matching accuracy and explore its potential in real application scenarios. 
Considering a surveillance environment, we can sense a pedestrian with a mobile phone by both videos for his/her visual appearance and wireless positioning for the carried phone. The video data captures the fine-grained appearance of pedestrians, but is vulnerable to occlusion. In contrast, the wireless positioning data is insensitive to occlusion, but suffers limited positioning accuracy.  In this work, we are dedicated to take advantage of the two heterogeneous but synchronous data for robust as well as accurate person re-identification. 

Since the video data and the wireless data are synchronous, the 2D visual trajectory of pedestrians in the video scene can be aligned to the wireless positioning trajectory. Such observation makes it feasible to bridge the visual data and the wireless positioning signal. 
When two video sequences of a pedestrian cannot be matched by visual data due to factors such as occlusion and posture changes, the consistency of the pedestrian motion trajectories captured by the two video sequences and the wireless positioning trajectory can help us match them correctly. Especially for longterm person re-identification (a few hours or a few days), mismatches caused by changes in pedestrian clothing can also be corrected by the appearance-invariance of the wireless signals. Therefore, the introduction of wireless signals can make up for some defects of visual signals.
Besides, through the wireless positioning trajectory, we can reduce the search space of person re-identification to regions near the trajectory.
Meanwhile, as a byproduct, some mobile phone numbers may be associated with the pedestrians' identity information such as name, ID number, \emph{etc.} Such cases make it possible to obtain the exact identities of the pedestrians when the faces are invisible or unclear. 


Based on the above discussion, we devise a recurrent context propagation module (RCPM) that contains a visual affinity update unit and a trajectory distance update unit. Through the alternating update of these two units, the visual information and positioning information propagate between the videos and the trajectories. The propagation and fusion of these two kinds of information make the algorithm more robust to noise such as occlusion and blur when searching pedestrians, and reduce the impact of positioning accuracy when matching wireless signals, \emph{i.e.,} the wireless positioning trajectories. After the learning, as shown in Fig.~\ref{fig:whole}, given a query video, we can retrieve the relevant videos and the corresponding wireless signal from the database. Considering the lack of dataset for this new task, we contribute a person re-identification dataset containing wireless positioning information for this task, which is named WP-ReID. Extensive experiments are carried out on our WP-ReID dataset to validate the effectiveness of the proposed method.

\section{Related Works}\label{sec:relatedWrok}

In this section, we first review the progress of person re-identification. Then, we briefly introduce the existing wireless positioning methods for mobile phones. Finally, we introduce some related cross-modality matching approaches.

\subsection{Person Re-identification} 

\textbf{Supervised person re-identification.} 
The occlusion, background clutter, and the dramatic variations in viewpoints and pedestrian postures bring lots of challenge to person re-identification. To solve these problems, many effective methods have been proposed, which are mainly divided into image-based person re-identification methods \cite{liu2016multi, chen2019abd, zhang2019densely, sun2019perceive, ro2019backbone, hou2019interaction, zhou2019omni} and video-based person re-identification methods \cite{mclaughlin2016recurrent, li2019multi, liu2019spatial, gu2019temporal, li2020relation}.

For image-based person re-identification, a lot of works are focused on designing feature extraction models that are robust to occlusion and blur. In~\cite{varior2016gated}, a gating function is proposed to compare the local patterns for an image pair starting from the mid-level and propagate more relevant features to the higher layers. 
In~\cite{li2018harmonious}, a novel Harmonious Attention CNN model is proposed for joint learning of soft pixel attention and hard regional attention along with simultaneous optimization of feature representations. 

For video-based person re-identification, many efforts are devoted to modeling the temporal clues along with the video frames. In~\cite{mclaughlin2016recurrent}, a novel recurrent DNN architecture is proposed to learn temporal representations and use the temporal-pooling to fuse the temporal features into a vector.
In~\cite{chung2017two}, the optical flow information is fed into the model to learn temporal features. 
In~\cite{liu2017quality, song2018region}, different branches are designed to predict quality scores for video frames. Liu \emph{et al.}~\cite{liu2019spatial} design a refining recurrent unit to suppress noise using the motion information and appearance representations.

\begin{figure*}[t]
  \centering
  \includegraphics[width=0.82\linewidth]{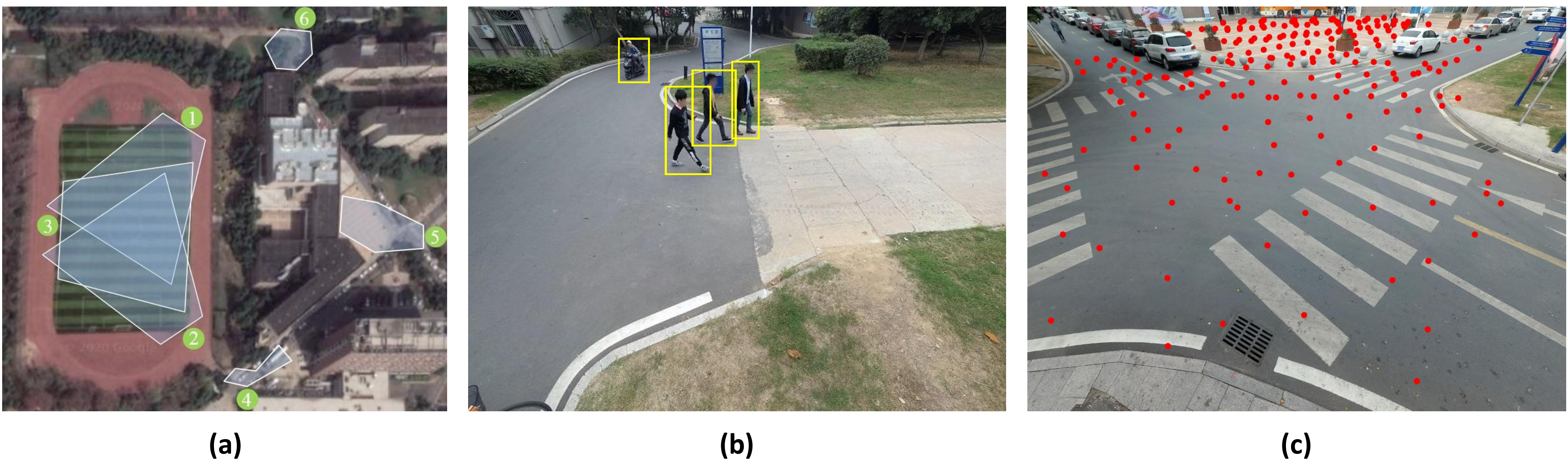}
  \caption{(a) The layout of the 6 cameras and the corresponding captured scenes. (b) One example of the captured scene. (c) The sampling points set for visual trajectory mapping.}
  \label{fig:dataset}
\end{figure*}

\textbf{Unsupervised Cross-domain Person Re-identification.} 
In the problem of cross-domain person re-identification \cite{han2018coteaching, yang2019patch, fan18unsupervisedreid, ge2020mutual, yu2019unsupervised}, with no supervision in the target domain, the goal is to transfer the identity discriminative representations from a labeled source domain to an unlabeled target domain. There are two main types of methods. The first one tries to align the source to the target domain by reducing the distribution gap. Han~\emph{et al}.~\cite{han2018coteaching} propose an image-to-image translation model, which maintains identities of source domain but transfer backgrounds of labeled source images to target camera style. 
However, this method only considers the inter-domain differences but ignores the intra-domain image style variations. 
In~\cite{fan18unsupervisedreid}, the intra-domain image style variations are reduced by learning with both unlabeled target images and their counterparts of the same identity but with different camera styles. 

For the second category of methods,  clustering-based adaptation has been widely explored. 
In ~\cite{fan18unsupervisedreid}, $k$-means is used to predict pseudo labels of unlabeled target images for model fine-tuning and use self-paced learning to ignore outliers. 
Ge~\emph{et al}.~\cite{ge2020mutual} use an online peer-teaching manner to conduct online robust soft pseudo labels.

\subsection{Wireless Positioning}
WiFi and cellular networks are principal ways to obtain personal location information.
For WiFi positioning, traditional techniques are usually based on the Received Signal Strength Indicator (RSSI) measurement~\cite{goswami2011wigem}. However, this strategy is usually unstable. To improve the stability, Channel State Information (CSI) measurement based methods~\cite{kotaru2015spotfi} are proposed to achieve better accuracy, with even decimeter level positioning. 
Localization methods~\cite{wigren2012lte,del2016performance} based on cellular networks have received  lots of attention from the first generation (1G) to the fifth generation (5G) wireless systems.  
Different methods such as trilateration, triangulation, and fingerprinting have been adopted to obtain position information. 
In~\cite{wang2019mobile}, by utilizing the beam information of the 5G wireless network, the authors achieve a positioning accuracy of about 12 meters.

\subsection{Cross-modality Matching}
Cross-modal matching has received extensive attention in many fields. In cross-modality matching between visual data and textual descriptions, many methods have been developed to generate captions for images \cite{karpathy2015deep, vinyals2015show} or videos \cite{venugopalan2014translating}. In cross-modality matching between visual data and radio signals, trajectory tracking of humans \cite{alahi2015rgb, papaioannou2015accurate}, face recognition via wireless signals \cite{lu2019autonomous} and person identification~\cite{korany2019xmodal} are exploited. 

In~\cite{korany2019xmodal}, a novel WiFi-video cross-modal gait-based person identification system (XModal-ID) is designed, which uses CSI magnitude measurements of a pair of off-the-shelf WiFi transceivers. XModal-ID uses the gait information to determine whether the WiFi measurements and a video belong to the same person. Different from XModal-ID~\cite{korany2019xmodal}, we judge whether a video and a wireless signal belong to the same person by the coincidence of the visual trajectory and wireless positioning trajectory. Meanwhile, the visual affinity is also considered, which greatly compensates for the limited positioning accuracy problem of wireless data and makes our system more reliable.

\section{WP-ReID Dataset}\label{sec:ourdataset}
 Since the existing person re-identification datasets do not contain wireless positioning information, we contribute one named Wireless Positioning Person Re-identification Dataset (WP-ReID).
WP-ReID contains not only visual data (video sequences of pedestrians) but also wireless signals (wireless positioning trajectories of the mobile phones carried by pedestrians). Based on WP-ReID, we can explore the potential of wireless signals together with video data on person re-identification and verify the effectiveness of different methods. In the following, we will briefly introduce the collection and setting of WP-ReID dataset.

\textbf{Video capture.} As shown in Fig.~\ref{fig:dataset}(a), we use 6 time-synchronized cameras with 4K ($4000 \times 3008$) resolution, of which three capture the playground from different directions, and the other three monitor the three gates of the playground. As the scenes captured by the surveillance cameras may overlap in real scenes (\emph{e.g.}, squares, and traffic intersections), the captured scenes of 3 of our cameras also overlap, which makes WP-ReID dataset similar to the real scenarios. Each camera collects 4,140 frames continuously at 6 fps. The final dataset contains video sequences of 79 pedestrians, 41 of whom are captured by at least two cameras. 

\textbf{Data labeling.} We manually initialize the pedestrian bounding boxes every 12 frames and generate the bounding boxes with the SiamRPN tracker~\cite{li2018high}  for the remaining 11 frames. We finally collect 868 video sequences of 79 pedestrians, with a total of 106,578 frames.

\textbf{Wireless positioning records.} 
In practice, we can proactively obtain the location information of mobile devices from WiFi or cellular networks, which, however, is beyond the scope of this work. 
Therefore, instead of estimating the mobile positioning information, we obtain it directly by a GPS software on the mobile phone, with a positioning error range of about 10 meters in our final dataset.
Such error range is close to that of existing positioning methods~\cite{kotaru2015spotfi,wang2019mobile} via WiFi or 5G networks.
Finally, the wireless signals of 29 pedestrians are recorded and these 29 pedestrians are captured by at least two cameras.

The wireless positioning trajectories (wireless signals) of 29 pedestrians are denoted as $\mathbf{T}^\mathrm{w} = \{\mathbf{T}^\mathrm{w}_m\}_{m=1}^M$ and $M = 29$. 
Each trajectory $\mathbf{T}^\mathrm{w}_m = \{\mathbf{T}^\mathrm{w}_{m, q}\}_{q=1}^{L^\mathrm{w}_m}$ is a set of two-dimensional coordinate points, which records the change of the position of the pedestrian every second. $L^\mathrm{w}_m$ is the total number of coordinate points. 

Each coordinate point $\mathbf{T}^\mathrm{w}_{m, q}$ in $\mathbf{T}^\mathrm{w}_m$ has a corresponding timestamp $\mathbf{H}^\mathrm{w}_{m, q}$ to indicate the time when it is recorded.
$\mathbf{T}^\mathrm{w}_m$ has the same identity label as the video data of the mobile phone holder.
Due to the wireless signal strength problem, the recording of some wireless positioning trajectories may be interrupted for a few seconds, which also makes our dataset closer to the real scenario.

\textbf{Visual trajectory mapping.} 
Video sequences and wireless positioning from the mobile phone both capture pedestrians' movements, but the two types of information are in different domains and cannot be directly linked. Therefore, we need a way to map these two kinds of information to the same domain.

\begin{table}[t]
\caption{The data splits of WP-ReID dataset.}
\label{tab:split}
\begin{center}
\begin{tabular}{l|cc|cc}
    \hline
    \multirow{2}{*}{Setting}  & \multicolumn{2}{c|}{Person Re-identification} & \multicolumn{2}{c}{Signal Matching} \\
    \cline{2-5}
    & \#ID & \#Video & \#ID & \#Data \\
    \hline
     Query & 41 & 201 & 29 & 164 videos\\
     Gallery & 79 & 868 & 29 & 29 signals\\
    \hline
  \end{tabular}
\end{center}
\end{table}

In an image captured by a camera, each pixel corresponds to a location in the real world, with a fixed latitude and longitude. This leads to a natural mapping between the pixel coordinate and the world coordinate. As shown in Fig.~\ref{fig:dataset}(c), we collect the latitude and longitude of some positions in the scene captured by the camera and further estimate the latitude and longitude of all pixels by interpolation. It is worth noting that for each surveillance camera, we only need to label the latitude and longitude information once, so this is once and for all.

Given a video frame, we consider that the latitude and longitude of the center point at the bottom of a pedestrian's bounding box (the position of the foot) is the location of the pedestrian. After being filtered by Kalman filter, we get the visual trajectory $\mathbf{T}^\mathrm{v}_i = \{\mathbf{T}^\mathrm{v}_{i, h}\}_{h=1}^{L^\mathrm{v}_i}$ of the pedestrian in the world coordinate system through $\mathbf{V}_i$. $L^\mathrm{v}_i$ is the total number of coordinate points. By recording the time when each frame is taken, we get the timestamp $\mathbf{H}^\mathrm{v}_{i, h}$ of each coordinate point $\mathbf{T}^\mathrm{v}_{i,h}$ in $\mathbf{T}^\mathrm{v}_i$.
Because the wireless signal recording and the camera both use the local time, their time is synchronized.  

We use $\mathbf{Q}_{i,m} $ to denote the coordinate point pairs with the same timestamp for $\mathbf{T}^\mathrm{v}_i$ and $\mathbf{T}^\mathrm{w}_m$, which is defined as
\begin{equation}
 \mathbf{Q}_{i,m} = \{(\mathbf{T}^\mathrm{v}_{i,h}\,, \mathbf{T}^\mathrm{w}_{m,q}) \, | \, \mathbf{H}^\mathrm{v}_{i,h} = \mathbf{H}^\mathrm{w}_{m,q} \,, h \in [1, L^\mathrm{v}_i] \,, q \in [1, L^\mathrm{w}_m]\} \,.
 \end{equation}
Then, the initial distance between the visual trajectory $\mathbf{T}^\mathrm{v}_i$ of video sequence $\mathbf{V}_i$ and \(m^{th}\) wireless trajectory $\mathbf{T}^\mathrm{w}_m$ is defined as 
\begin{equation}
\label{equ:dist:vw}
\mathbf{D}^0_{i,m} = \left\{
  \begin{array}{l}
  \frac{ \sum_{ (\mathbf{T}^\mathrm{v}_{i,h}\,, \mathbf{T}^\mathrm{w}_{m,q}) \, \in  \,  \mathbf{Q}_{i,m} }{ \mathrm{Euc}(\mathbf{T}^\mathrm{v}_{i,h}\,, \mathbf{T}^\mathrm{w}_{m,q}) } }{\lvert \mathbf{Q}_{i,m} \rvert}   \quad   \mathrm{if} \, \mathbf{Q}_{i,m} \notin \varnothing \,,\\
\infty \qquad \qquad \qquad  \qquad \qquad \qquad \mathrm{if} \, \mathbf{Q}_{i,m} \in \varnothing \,,
\end{array} \right.
\end{equation}
where $\mathrm{Euc}(\cdot, \cdot)$ calculates the Euclidean distance between two points. As the wireless positioning data is recorded once per second and the video data is recorded at 6 frames per second. Therefore, only the visual trajectory data recorded on the time of integer seconds will participate in the distance calculation. In WP-ReID dataset, the average distance between the visual trajectory and the corresponding wireless trajectory with the same identity is 9.1 meters.

\textbf{Evaluation protocol.} 
WP-ReID dataset is only used as a test dataset and does not contain any training data. For 41 pedestrians captured by at least two cameras, we
randomly select a video sequence for each pedestrian under each camera as the query video sequence.
During the test, given a video sequence of a pedestrian of interest, we have two subtasks. One is to find the video sequences of this person under other cameras like the traditional person re-identification. Similar to existing person re-identification datasets \cite{zheng2016mars, wu2018exploit}, we take the Cumulated Matching Characteristics (CMC) table and mean Average Precision (mAP) as the evaluation metric. The other subtask is to find the corresponding wireless signal of this pedestrian, which utilize the CMC table as the evaluation metric. 


As shown in Table~\ref{tab:split}, the final person re-identification query set contains 201 video sequences. Among these 201 video sequences, 164 video sequences with corresponding wireless signals are used to evaluate the performance of signal matching. The entire dataset containing 868 video sequences of 79 pedestrians constitutes the gallery set of person re-identification. The 29 wireless signals $\mathbf{T}^\mathrm{w} = \{\mathbf{T}^\mathrm{w}_m\}_{m=1}^M$ (wireless trajectories) constitutes the gallery set of signal matching.

\section{Our Method}\label{sec:ourMethod}

\subsection{Overview}
Our database contains not only video data, but also wireless signal data (wireless trajectories). The introduction of wireless signals can make up for the defects of the sensitivity of visual data to factors such as occlusion, blur, posture change, and clothing change.
Our problem involves two subtasks.
Given a query video sequence of a pedestrian, we not only need to find out the videos belonging to this person from the database but also identify which wireless signal belongs to this person from multiple wireless signals.
After finding the wireless signal of the pedestrian (pedestrian's mobile phone), 
we can narrow down the search range of this person to the vicinity of the trajectory, and have the opportunity to obtain the exact identity (\emph{e.g.}, name and ID number) of the pedestrian.

To associate the pedestrian's video sequences with its corresponding wireless signal, 
one approach is to consider the wireless trajectory closest to the visual trajectory of the query video sequence as the corresponding wireless trajectory. However, due to the limited positioning accuracy, when two pedestrians are close to each other, the wireless trajectory closest to the visual trajectory may not be the correct one. This limits the matching accuracy of this method.

\begin{figure*}[ht]
  \centering
  \includegraphics[width=0.67\linewidth]{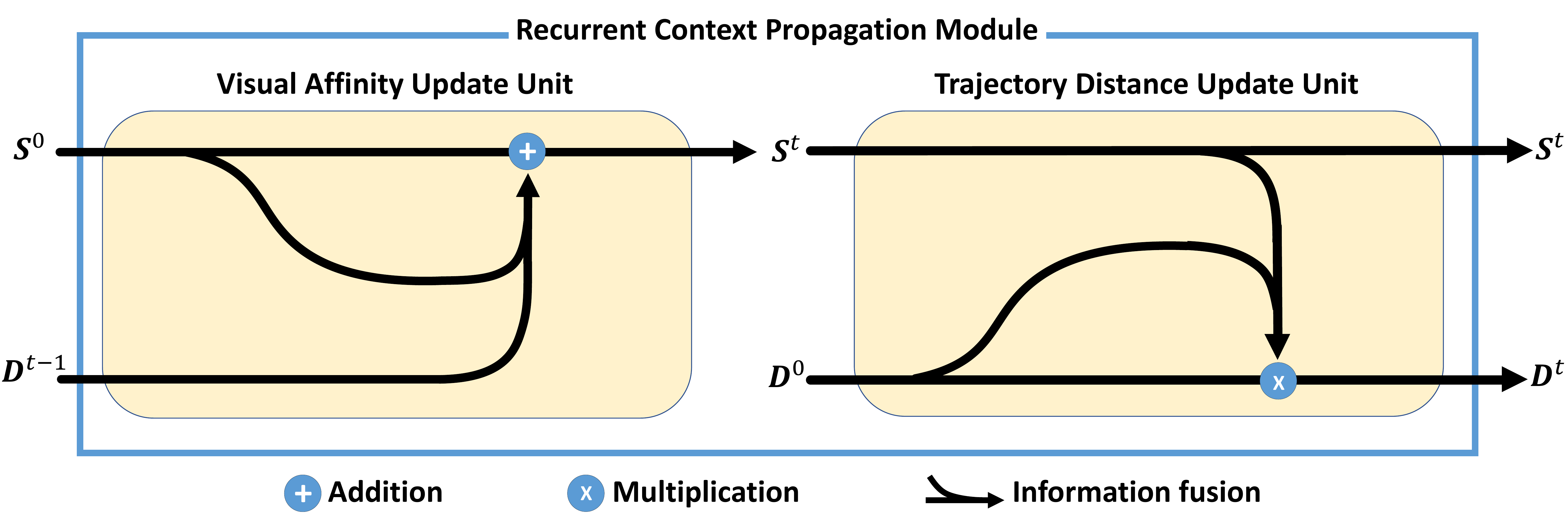}
  \caption{The architecture of the recurrent context propagation module, which contains a visual affinity update unit and a trajectory distance update unit  . The propagation and fusion of visual affinity information and trajectory distance information are displayed. $\mathbf{S}^{t}$ and $\mathbf{D}^{t}$ are the visual affinity and trajectory distance after the \({t}^{\mathrm{th}}\) update, respectively. The information fusion represents different ways to fuse information of different modalities.}
  \label{fig:rapm}
\end{figure*}

To handle the problem mentioned above, as shown in Fig.~\ref{fig:rapm}, we propose a recurrent context propagation module, which propagates and fuses visual affinity and trajectory distance  between videos and trajectories. 
By exploring the complementarity of data in different modalities, the two sub-tasks promote each other to obtain higher matching accuracy. 

\subsection{Visual Affinity and Trajectory Distance}
For the $N$ video sequences $\mathbf{V} = \{\mathbf{V}_i\}_{i=1}^N$ in the database, we measure the visual distance between $\mathbf{V}_i$ and $\mathbf{V}_j$ by 
\begin{equation}
\mathbf{F}_{i,j} = f(\mathbf{V}_i, \mathbf{V}_j) \,,
\end{equation}
where $f(\cdot, \cdot)$ is a distance measurement function (\emph{e.g.}, Euclidean distance of the features).
Then, we define $\mathbf{S}^0_{i,j}$ as the initial visual affinity between
$\mathbf{V}_i$ and $\mathbf{V}_j$, which is measured as:
\begin{equation}
\label{equ:def:s}
  \mathbf{S}^0_{i,j} = 1 - \frac{\mathbf{F}_{i,j} - [\mathbf{F}_i]_{\mathrm{min}}}{[\mathbf{F}_i]_{\mathrm{max}} - [\mathbf{F}_i]_{\mathrm{min}}} \,,
\end{equation}
where $\mathbf{F}_i \in {\mathbb{R}}^{N}$ denotes the distance between $\mathbf{V}_i$ and other video sequences. $[\cdot]_{\mathrm{max}}$ and $[\cdot]_{\mathrm{min}}$ return the maximum and minimum values in the matrix, respectively. With Eq.~\ref{equ:def:s}, we normalize the distance matrix to a range between 0 and 1 by linear mapping and get the the initial visual affinity $\mathbf{S}^0_{i,j}$.
The more similar the two video sequences are, the closer their visual affinity will be to 1. Therefore, the visual affinity $\mathbf{S} \in {\mathbb{R}}^{N \times N}$ models the visual similarity between video sequences in the database.

Meanwhile, given the visual trajectories $\mathbf{T}^\mathrm{v} = \{\mathbf{T}^\mathrm{v}_i\}_{i=1}^N$ of the video sequences $\mathbf{V} = \{\mathbf{V}_i\}_{i=1}^N$ and the $M$ wireless signals (wireless trajectories) $\mathbf{T}^\mathrm{w} = \{\mathbf{T}^\mathrm{w}_m\}_{m=1}^M$, the initial trajectory distance $\mathbf{D}^0 \in {\mathbb{R}}^{N \times M}$ is defined by Eq.~\ref{equ:dist:vw}. The smaller the value of $\mathbf{D}^0_{i,m}$ is, the more the two trajectories match, and the more likely the two trajectories belong to the same person. 

\subsection{Visual Affinity Update Unit}
We take the wireless trajectories as anchors to measure the possibility that two visual trajectories come from the same pedestrian. 
As shown in Fig.~\ref{fig:whole}, when two video sequences belong to the same person, their visual trajectories should be close to the corresponding wireless trajectory. Therefore, when both the visual trajectories of two video sequences are close to a wireless trajectory, the two video sequences are likely to belong to the same person. The magnitude of the likelihood can be measured by their average distance from this trajectory.

Based on the above observation, we designed a visual affinity update unit. Given two video sequences $\mathbf{V}_i$ and $\mathbf{V}_j$, we obtain $ \mathbf{D}^{t-1}_i = [\mathbf{D}^{t-1}_{i,1}, \mathbf{D}^{t-1}_{i,2}, \cdots, \mathbf{D}^{t-1}_{i,M}]$ and $ \mathbf{D}^{t-1}_j = [\mathbf{D}^{t-1}_{j,1}, \mathbf{D}^{t-1}_{j,2}, \cdots, \mathbf{D}^{t-1}_{j,M}]$, which measures the distances between their visual trajectories and all $M$ wireless trajectories. We define $ \mathbf{\hat{D}}^{t-1}_{i,j} = {[ \frac{\mathbf{D}^{t-1}_i + \mathbf{D}^{t-1}_j}{2} ]}_\mathrm{min}$ to get the smallest average distance from these two video sequences.

When $\mathbf{\hat{D}}^{t-1}_{i,j}$ is small, the two video sequences may belong to the same person. 
However, when this value is large, we cannot make a judgment that they do not belong to the same
person. This is because we may not be able to find a close trajectory for the two videos of one person that does not have a corresponding wireless signal. 
Therefore we define a distance threshold $\sigma$ to help decide whether to update the trajectory distance. If $\mathbf{\hat{D}}^{t-1}_{i,j}$ is less than $\sigma$, the visual affinity between $\mathbf{V}_i$ and $\mathbf{V}_j$ will be updated, otherwise it is fixed. Based on the above disucssion, the visual affinity update unit is defined as
\begin{equation}
\label{equ:u:va}
  \mathbf{S}^t_{i,j} = \left\{
  \begin{array}{l}
\mathbf{S}^{0}_{i,j} \qquad \qquad \qquad \quad \mathrm{if} \; \mathbf{\hat{D}}^{t-1}_{i,j} >  \sigma \;  \mathrm{or} \;  i = j \,,\\
\mathbf{S}^0_{i,j} \times 0.5 + \left( 1 - \frac{\mathbf{\hat{D}}^{t-1}_{i,j}}{\sigma} \right) \times 0.5 \quad   \mathrm{otherwise} \,,
\end{array} \right.
\end{equation}
where $\frac{\mathbf{\hat{D}}^{t-1}_{i,j}}{\sigma}$ is to normalize $\mathbf{\hat{D}}^{t-1}_{i,j}$ into the range between 0 and 1. To prevent error accumulation, 
the visual affinity to be updated in the visual affinity update unit is the initial affinity $\mathbf{S}^0$ instead of $\mathbf{S}^{t-1}$.

In the visual affinity update unit, we introduce the trajectory distance to update visual affinity. The similarity of the two video sequences not only depends on their visual similarity but also takes into account the coincidence with the wireless trajectory. This can reduce the impact of video noise and is useful for matching some videos with severe occlusion and blur.

\subsection{Trajectory Distance Update Unit}
Different visual trajectories of the same person should all coincide with the corresponding wireless trajectory. 
When the visual trajectory $\mathbf{T}^\mathrm{v}_i$ of a video sequence $\mathbf{V}_i$ matches a wireless trajectory $\mathbf{T}^\mathrm{w}_m$, the visual trajectories of the video sequences with higher visual affinity to $\mathbf{V}_i$ should also be close to $\mathbf{T}^\mathrm{w}_m$, because the high affinity indicates that these videos and $\mathbf{V}_i$ are likely to belong to the same person. We use the $K$-nearest neighbors $\psi(i, t, K)$ to denote the set of video sequences that are similar to $\mathbf{V}_i$ in \(t^{th}\) update, which is obtained by sorting $\mathbf{S}^t_{i}$ in descending order  and getting the top-$K$ video sequences. It is worth noting that the video sequence $\mathbf{V}_i$ itself is also in $\psi(i, t, K)$.

We consider using the distance values $\{\mathbf{D}^0_{k,m} | \, k \in \psi(i, t, K)\}$ of these video sequences $\{\mathbf{V}_k | \, k \in \psi(i, t, K)\}$ from the wireless trajectory $\mathbf{T}^\mathrm{w}_m$ to update the trajectory distance $\mathbf{D}_{i,m}$ of $\mathbf{V}_i$, but this update is not simply averaging these distance values. For a video sequence with a higher visual affinity with $\mathbf{V}_i$, it is more likely to belong to the same person with $\mathbf{V}_i$. At the same time, its distance from $\mathbf{T}^\mathrm{w}_m$ has more reference value, which means that its weight should be higher. Therefore, we consider using the affinity value as the weighted value of distance. Then, we design a trajectory distance update unit, which updates the trajectory distance $\mathbf{D}$ with the information from visual affinity $\mathbf{S}$ as follows, 
\begin{align}
\label{equ:u:ta}
  \mathbf{D}^t_{i,m} & = \frac{ \sum_{k \in \psi(i, t, K)}{\mathbf{D}^0_{k,m}} \times  \mathbf{S}^{t}_{i,k} \times I(\mathbf{D}^0_{k,m}) }{ \sum_{k \in \psi(i, t, K)}{\mathbf{S}^t_{i,k}  \times I(\mathbf{D}^0_{k,m})}} \,, 
\end{align}
where $I(\mathbf{D}^0_{k,m}) = 1$ when $\mathbf{D}^0_{k,m} \neq \infty$, otherwise  $I(\mathbf{D}^0_{k,m}) = 0$. The introduction of $I(\cdot)$ is to eliminate the influence of neighboring video sequences that have no overlapping time periods with this wireless signal. 
If all  $\{I(\mathbf{D}^0_{k,m}) | \, k \in \psi(i, t, K)\}$ are 0, the trajectory distance $\mathbf{D}^t_{i,m}$ is directly set as $\mathbf{D}^0_{i,m}$.
To prevent error accumulation, the inputs of the unit are visual affinity $\mathbf{S}^{t}$ and the initial trajectory distance $\mathbf{D}^0$ rather than $\mathbf{S}^{t}$ and $\mathbf{D}^{t-1}$. 

In the trajectory distance update unit, we use visual affinity as the weighting value. The updated distance between a video sequence (visual trajectory) and a wireless trajectory is the weighted average result of the distance between its neighboring video sequences and this wireless trajectory.
Through the propagation of information from visual affinity to trajectory distance, the distance between a visual trajectory and a wireless trajectory will not only depend on itself, but will also be affected by similar video sequences. This makes our method insensitive to the positioning accuracy when matching wireless signals.

\subsection{Recurrent Context Propagation Module}

As shown in Fig.~\ref{fig:rapm}, after combining the visual affinity update unit and the trajectory distance update unit, we obtain the recurrent context propagation module (RCPM). Given the initial visual affinity $\mathbf{S}^0$ and trajectory distance $\mathbf{D}^0$, the visual affinity and trajectory distance will propagate and merge through the multiple iterative updates of RCPM. The final visual affinity includes not only visual information but also wireless positioning information. Similarly, the final trajectory distance also contains information on the two modalities.

In summary, in order to handle the two tasks of person re-identification and signal matching, our method includes two stages: feature extraction and affinity update. First, we can train a feature extraction model through supervised or unsupervised methods. We merge the query video sequence $\mathbf{V}_{i}$ and the gallery video sequences into a database. After that, we use the feature extraction model to extract features and get the visual affinity $\mathbf{S}^0$ of this database. Then, by calculating the distance between the visual trajectories of the video sequences and the wireless trajectories, we obtain the initial trajectory distance $\mathbf{D}^0$. After being updated by RCPM, we get the updated visual affinity $\mathbf{S}^{t}$ and trajectory distance $\mathbf{D}^t$. In the end, we will sort $\mathbf{S}^{t}_i$ and $\mathbf{D}^{t}_i$ to get the pedestrian video sequences and wireless signal most similar to the query video sequence $\mathbf{V}_{i}$.

\textbf{How to speed up person re-identification in large-scale databases?} In a large video surveillance network, we can first construct a small database using the videos captured by the cameras within a certain range around the location of the query video. After finding the person's wireless signal through our proposed algorithm, we can only retrieve the videos captured by the cameras near the wireless trajectory. Therefore, unlike the existing methods, our strategy avoids searching the videos in the entire surveillance network directly, which is of great significance for person re-identification in the application to real scenes.

\section{Experiment}\label{sec:experiments}

\subsection{Datasets and Protocols}
\label{exp:datasets}
To help train the feature extraction models, we select two large publicly available video-based person re-identification datasets including MARS \cite{zheng2016mars} dataset and DukeMTMC-VideoReID \cite{wu2018exploit} dataset. 

MARS \cite{zheng2016mars} consists of 1,261 identities and 20,715 video sequences under 6 cameras. The bounding boxes are generated by classic detection and tracking algorithms. DukeMTMC-VideoReID \cite{wu2018exploit} is a subset of the tracking dataset DuKeMTMC \cite{ristani2016performance} and contains 4,832 tracklets from 1,812 identities. 

We follow the standard training and testing splits of MARS and DukeMTMC-VideoReID. Only the training set is used as the source dataset to help train the feature extraction model. In all experiments, we take the standard Cumulated Matching Characteristics (CMC) table and mean Average Precision (mAP) as the evaluation metric.

\begin{table*}[t]
\caption{The person re-identification results and signal matching results of different methods on WP-ReID dataset. Because of space constraints, we select four representative existing person re-identification methods to be combined with the proposed recurrent context propagation module (RCPM) and gave their performances. The default setting for RCPM is $\sigma = 74$ and 4 iterations. The optimal $K$ of different methods on different source dataset is also given in the table. \emph{SM-Baseline} is the baseline method of signal matching task, which directly sorts the initial distance between the visual trajectories and the wireless trajectories and takes the nearest one as the correct wireless signal.}
\label{tab:traj:ablation}
\begin{center}
\begin{tabular}{l|c|cccc|ccc|c|cccc|ccc}
    \hline
    \multirow{3}{*}{Method}  & \multicolumn{8}{c|}{MARS $\to$ WP-ReID} & \multicolumn{8}{c}{DukeMTMC-reID $\to$ WP-ReID} \\
    \cline{2-17}
    & \multirow{2}{*}{$K$} & \multicolumn{4}{c|}{Person Re-identification} & \multicolumn{3}{c|}{Signal Matching} & \multirow{2}{*}{$K$} & \multicolumn{4}{c|}{Person Re-identification} & \multicolumn{3}{c}{Signal Matching}\\
    \cline{3-9}  \cline{11-17}
    & & mAP & R1 & R5 & R10 & R1 & R5 & R10 & & mAP & R1 & R5 & R10 & R1 & R5 & R10 \\
    \hline

    SM-Baseline    & - & - & - & - & - & 53.7 & 93.9 & 96.3    & - & - & - & - & - & 53.7 & \textbf{93.9} & 96.3      \\
    \hline
    M3D \cite{li2019multi}    & - & 11.5 & 30.3 & 49.3 & 61.7  & - & - & -    & - & 15.6 & 42.8 & 67.2 & 74.6 & - & - & -    \\
    I3D \cite{carreira2017quo}   & - & 14.6 & 35.8 & 56.7 & 66.7 & - & - & -    & - & 13.1 & 33.3 & 58.2 & 72.1 & - & - & -    \\
    P3D \cite{qiu2017learning}    & - & 11.1 & 29.9 & 46.8 & 55.7 & - & - & -    & - & 10.8 & 28.9 & 50.7 & 59.7 & - & - & -    \\
    HHL \cite{zhong2018generalizing}    & - & 23.3 & 50.2 & 73.1 & 78.1 & - & - & -    & - & 32.0 & 57.2  & 76.6 & 88.1 & - & - & -    \\
    \hline
    TKP \cite{gu2019temporal}    & - & 15.1 & 41.3 & 58.7 & 68.7 & - & - & -    & - & 26.1 & 57.2 & 73.6 & 80.1 & - & - & -      \\
    TKP \cite{gu2019temporal} + RCPM    & 4 & 32.5 & 57.7 & 72.1 & 77.1 & 64.6 & 84.7 & 95.1    & 6 & 47.5 & 67.2 & 76.6 & 83.6 & 73.8 & 91.5 & \textbf{97.0}     \\
    STMP \cite{liu2019spatial}    & - & 26.2 & 53.2 & 73.6 & 80.1 & - & - & -    & - & 36.8 & 64.2 & 80.6 & 86.1 & - & - & -     \\
    STMP \cite{liu2019spatial} + RCPM     & 5 & 46.2 & 68.2 & 75.1 & 80.6 & 73.2 & 89.0 & 95.1    & 5 & 60.4 & \textbf{78.1} & 80.6 & 83.5 & 84.2 & 93.3 & 94.5     \\
    SSG \cite{fu2019self}    & - & 24.7 & 60.2 & 70.1 & 75.1 & - & - & -    & - & 28.2 & 63.7 & 77.1 & 82.1 & - & - & -     \\
    SSG \cite{fu2019self} + RCPM    & 4 & 42.5 & 67.7 & 79.1 & \textbf{86.6} & 74.4 & \textbf{93.9} & \textbf{97.0}    & 4 & 47.8 & 72.1 & 79.6 & 85.1 & 76.2 & 92.7 & 95.7     \\
    MMT \cite{ge2020mutual}     & - & 32.8 & 64.2 & 79.1 & 82.6 & - & - & -    & - & 39.1 & 72.6 & \textbf{82.1} & 86.1 & - & - & -     \\
    MMT \cite{ge2020mutual} + RCPM    & 8 & \textbf{55.3} & \textbf{74.1} & \textbf{80.1} & 83.6 & \textbf{83.5} & 92.1 & 95.7    & 8 & \textbf{61.6}  & 77.1 & \textbf{82.1} & \textbf{86.6} & \textbf{89.6} & 93.3 & 96.3     \\
    
    \hline
  \end{tabular}
\end{center}
\end{table*}

\subsection{Implementation Details}
\label{exp:implement}

We select some supervised and unsupervised person re-identification algorithms to train the feature extractors. For supervised methods \cite{li2019multi, carreira2017quo, qiu2017learning, gu2019temporal, liu2019spatial}, we train the feature extractors directly on the source dataset and then test it on WP-ReID dataset. For the unsupervised methods \cite{zhong2018generalizing, fu2019self, ge2020mutual}, they not only need to train the networks on the source dataset but also need to perform unsupervised training again on the training set of the target dataset. Since our dataset has no training set, these methods are optimized directly on WP-ReID, but no labels are used.

As MMT \cite{ge2020mutual} takes too long to train the model on MARS and DukeMTMC-VideoReID, we downsample all video sequences to 1/16 of their original length to speed up training. For other supervised and unsupervised person re-identification methods, we use the default training settings and parameters.

\subsection{Ablation Study}
\label{exp:ablation}

\begin{figure*}[t]
  \centering
  \includegraphics[width=0.86\linewidth]{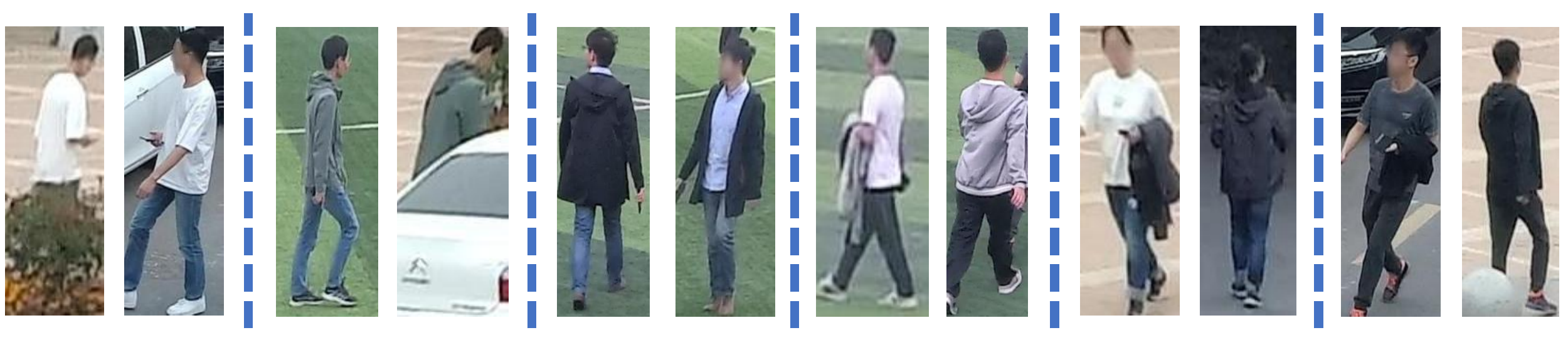}
  \caption{Examples that the methods based on visual data cannot correctly discriminate due to occlusion and the change of clothes and viewpoints. These examples are correctly identified with the aid of wireless signals. The images in the same block share the same person identity.}
  \label{fig:person}
\end{figure*}

\textbf{Results of different methods on WP-ReID.}
As shown in Table~\ref{tab:traj:ablation}, we show the person re-identification performances of some existing methods on WP-ReID dataset, including supervised and unsupervised methods.
The source datasets are MARS and DukeMTMC-VideoReID.
The performance of these methods is not satisfactory, which shows that the WP-ReID dataset is very challenging. This is because there are lots of occlusion and blur in WP-ReID dataset. Meanwhile, as shown in Fig.~\ref{fig:person}, we observe that some pedestrians take off their jackets when walking, which also introduces a great challenge for the existing methods to match pedestrians due to the dependence on the appearance representations.  

Occlusions, change of viewpoints, and taking off the jackets are common phenomena in real scenes, but these problems are difficult to solve with existing person re-identification methods. Existing methods only consider the visual information, and the performance is poor when the visual data is unreliable. Therefore, we need to introduce wireless positioning information to reduce the influence of noises by the complementary information between data in different modalities.

\textbf{The effectiveness of RCPM.}
We show the performance comparison of some existing methods with or without the proposed recurrent context propagation module (RCPM) in Table~\ref{tab:traj:ablation}. 
Person re-identification and signal matching are two independent tasks, which rely on visual data and positioning data, respectively. But when RCPM is adopted, the model will consider the data of two modalities at the same time, and the performance of the model on one task will be affected by the data of another task.
We set $\sigma = 74$ and the number of iterations to 4 for RCPM. Different models have different discrimination capabilities for pedestrians, so the optimal values of $K$ are also different when defining the range of similar video sequences of a video sequence.
The optimal values of $K$ of different models are listed in the table. 

It can be seen that when RCPM is adopted, the performances of person re-identification is greatly improved. This shows that the introduction of wireless signals can help improve the performance of person re-identification. As shown in Fig.~\ref{fig:person}, these examples that cannot be identified by existing methods due to occlusion and changes in clothing can be correctly identified by our method.
Through the update of RCPM, the similarity between two video sequences depends not only on their visual information but also on the coincidence with the wireless trajectory. If one person's distinguishing parts are obscured, the feature of his video sequence may be very different from the feature of another video sequence of this person, but the consistency of the visual trajectories of the two video sequences with the wireless trajectory can make our method still correctly identify them.

\begin{figure*}[ht]
  \centering
\subfloat[]{
\begin{minipage}[t]{0.28\linewidth}
\centering
\includegraphics[width=0.95\linewidth]{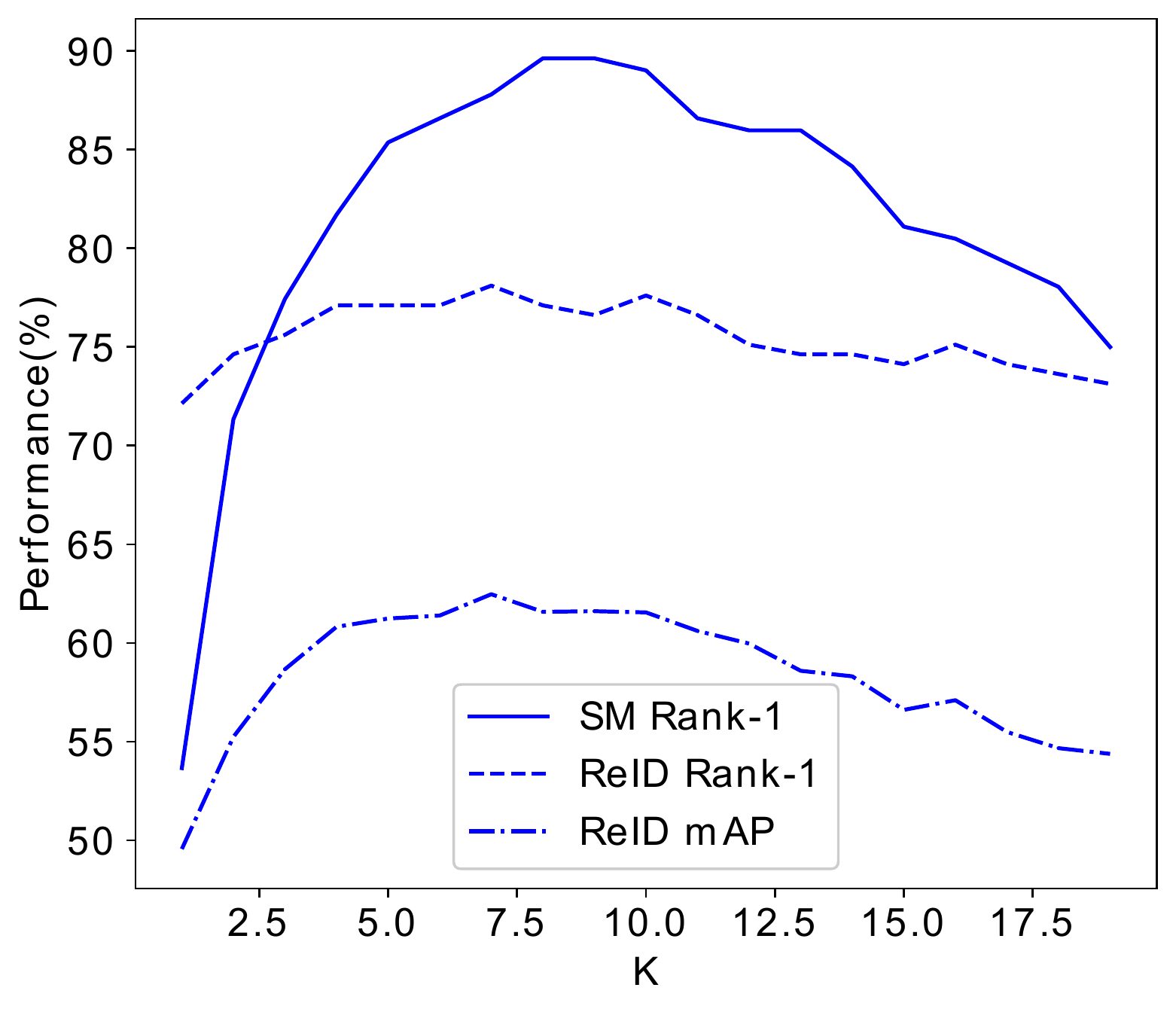}
\end{minipage}%
}%
\subfloat[]{
\begin{minipage}[t]{0.28\linewidth}
\centering
\includegraphics[width=0.95\linewidth]{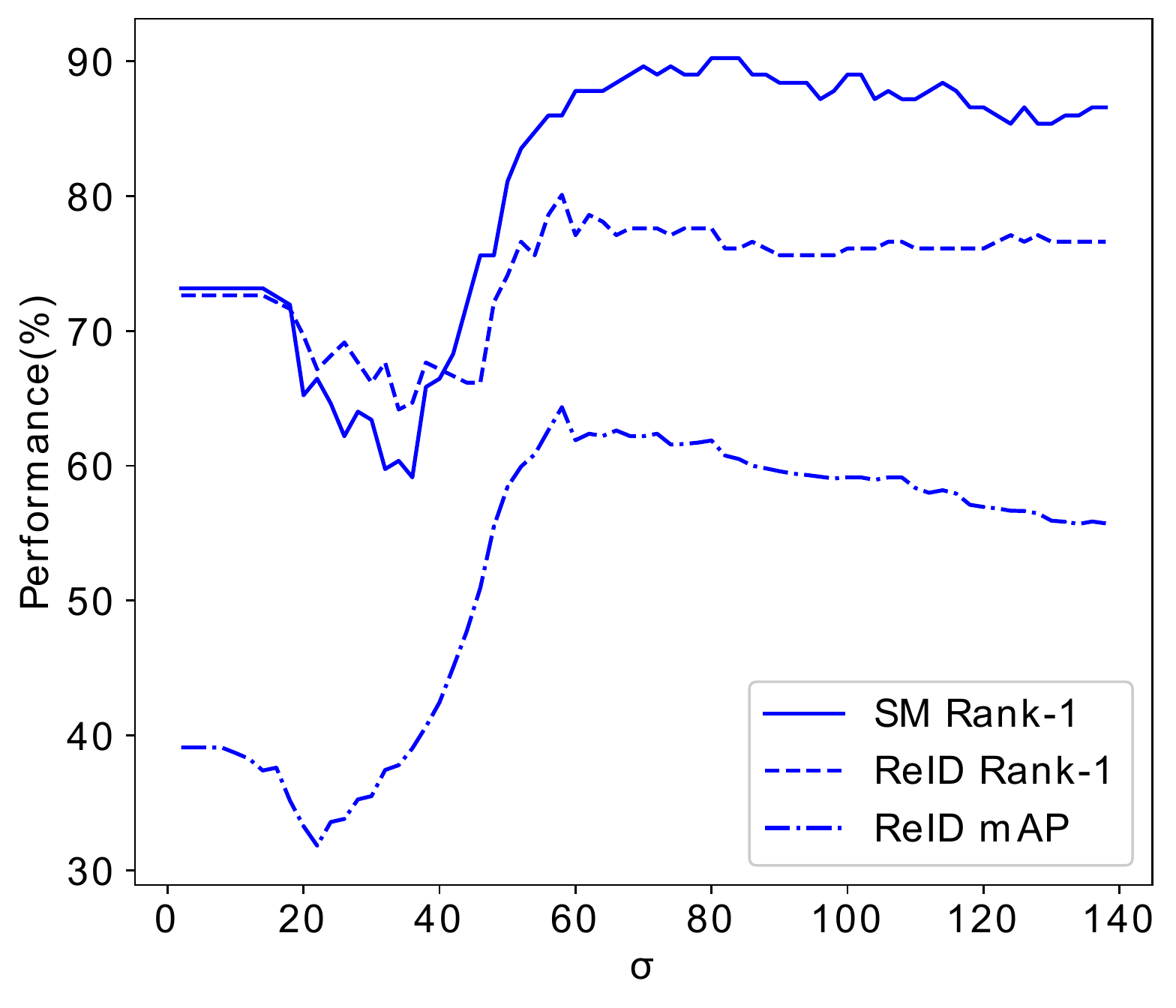}
\end{minipage}%
}%
\subfloat[]{
\begin{minipage}[t]{0.345\linewidth}
\centering
\includegraphics[width=0.95\linewidth]{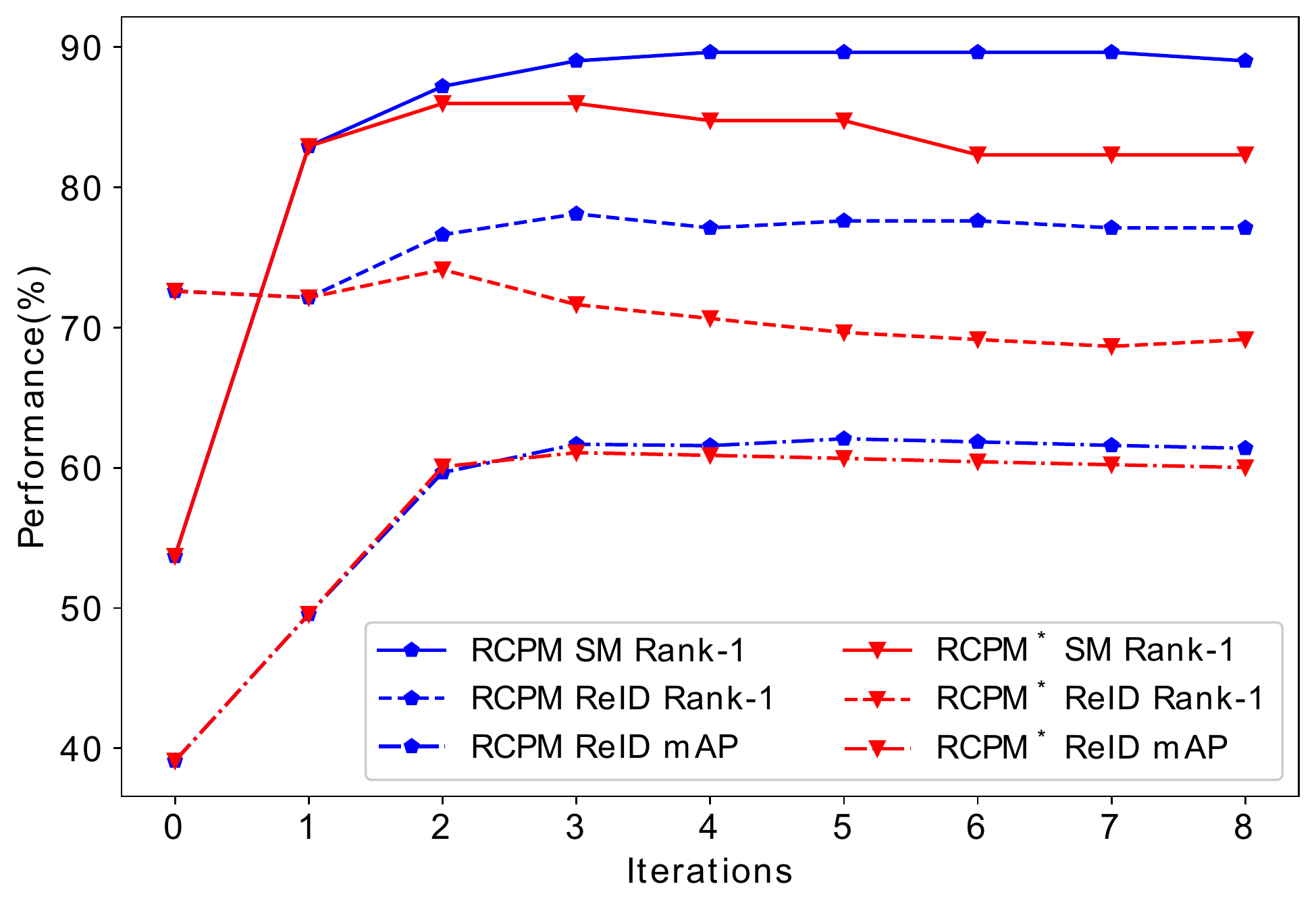}
\end{minipage}
}%
\centering
\caption{The ablation study of the recurrent context propagation module (RCPM). The model we use to extract features and calculate the initial visual affinity is trained by the method MMT \cite{ge2020mutual} that takes the DukeMTMC-VideoReID as the source dataset and WP-ReID as the target dataset. The default setting for RCPM is $K = 8, \sigma = 74$ and 4 iterations. The \emph{SM} in the figure denotes signal matching.  (a) The performance of RCPM at different $K$. (b) The performance of RCPM at different $\sigma$. (c) The performance of RCPM at different iterations. The \emph{RCPM} in the figure denotes the raw RCPM. $\mathbf{RCPM}^*$ is a modified version of RCPM. In $\mathbf{RCPM}^*$, the inputs of the visual affinity update unit are $\mathbf{S}^{t-1}$ and $\mathbf{D}^{t-1}$ instead of $\mathbf{S}^{0}$ and $\mathbf{D}^{t-1}$. Meanwhile, the trajectory distance update unit takes the $\mathbf{S}^{t}$ and $\mathbf{D}^{t-1}$ as inputs instead of $\mathbf{S}^{t}$ and $\mathbf{D}^{0}$.
}
\label{fig:rapm:ablation}
\end{figure*}

In Table~\ref{tab:traj:ablation}, the baseline method of signal matching is \emph{SM-Baseline}, which directly sorts the initial distance between the visual trajectories and the wireless trajectories and takes the nearest one as the correct wireless signal. It can be seen that the rank-1 accuracy of the baseline method is only $53.7\%$. This is because, in WP-ReID, pedestrians walk unconstrained, and some pedestrians walk with each other. When pedestrians walk close and the distances between them are less than the positioning error range of the positioning method, the wireless trajectories of different pedestrians cannot be distinguished directly based on the trajectory distances. However, it can be seen from Table~\ref{tab:traj:ablation} that the method \emph{MMT + RCPM} using DukeMTMC-reID as source dataset achieves a $35.9\%$ signal matching rank-1 accuracy gain compared with \emph{SM-Baseline}. The rank-1 accuracy of signal matching is greatly improved through the update of our method RCPM. This is because in our method, the distance between the visual trajectory of a video sequence and a wireless trajectory is not only related to its trajectory distance but also the distances between the visual trajectories of the video sequences similar to this video and the wireless trajectory. This operation enables our method to mitigate the effects of positioning accuracy in signal matching through visual information. 

As shown in Table~\ref{tab:traj:ablation}, the person re-identification mAP accuracy of TKP \cite{gu2019temporal} using MARS dataset as source dataset and WP-ReID as target dataset is only $15.1\%$. But RCPM still boosts the signal matching performance of \emph{TKP + RCPM}, which shows that RCPM can still work when the performance of the person re-identification model is poor. Meanwhile, we notice that when the person re-identification model has a better ability to distinguish pedestrians, the algorithm obtains higher signal matching performance after using RCPM. The MMT \cite{ge2020mutual} trained from the DukeMTMC-reID dataset has $72.6\%$ rank-1 accuracy of person re-identification and the rank-1 accuracy of signal matching accuracy of the baseline method is only $53.7\%$, RCPM improves the accuracy of signal matching to $89.0\%$. This shows that if there is a model with better pedestrian distinguish ability and a positioning strategy with higher positioning accuracy, RCPM can improve the matching accuracy to closer to $100.0\%$. So it is very feasible to apply our method to the practical scenario to find the corresponding wireless signals.

\textbf{The ablation study of RCPM.}
As shown in Fig.~\ref{fig:rapm:ablation}, we show the performance changes of the RCPM using DukeMTMC-reID as the source dataset when different parameters change. As $K$ increases, the model's performance increases as more information from similar video sequences is taken into account. 
However, as $K$ continues to increase, the performance of the model decreases. This is because when $K$ is too large, video sequences in the $K$-nearest neighbors will be doped with too many video sequences of different identities and too much noise is introduced into the affinity update. 
As the threshold $\sigma$ increases, the overall performances of the model increase first and then decrease slowly. When $\sigma = 74$ , the comprehensive performance reaches the highest. 

As shown in Fig.~\ref{fig:rapm:ablation}(c), for the raw RCPM, as the number of iterations increases, the performance of the model improves first and then stabilizes after 4 iterations. As mentioned in Section \ref{sec:ourMethod}, in order to prevent error accumulation, the information to be updated in each update unit is the initial affinity. In Fig.~\ref{fig:rapm:ablation}(c), we also display the result of the modified version of RCPM, \emph{i.e.,} $\mathrm{RCPM}^*$. In $\mathrm{RCPM}^*$, the inputs of the visual affinity update unit are $\mathbf{S}^{t-1}$ and $\mathbf{D}^{t-1}$ instead of $\mathbf{S}^{0}$ and $\mathbf{D}^{t-1}$. Meanwhile, the trajectory distance update unit takes the $\mathbf{S}^{t}$ and $\mathbf{D}^{t-1}$ as inputs instead of $\mathbf{S}^{t}$ and $\mathbf{D}^{0}$. The experimental results show that this $\mathrm{RCPM}^*$ has a significant performance degradation after several iterations. This denotes that our method of preventing error accumulation is effective and necessary.

\section{Conclusion}\label{sec:conclusion}

In this paper, we study the problem of person re-identification with two heterogeneous data, \emph{i.e.,} video and wireless mobile positioning signal. Given a video sequence of a pedestrian, we need to find out the videos belonging to this person and find the corresponding wireless signal. To this end, we contribute  a person re-identification dataset WP-ReID containing wireless signals.   Meanwhile, we propose a recurrent context propagation module that enables visual information and wireless positioning information to propagate and fuse and makes these two sub-tasks mutually improve performance. The introduction of wireless signals can make up for the defects of visual data sensitive to factors such as occlusion, blur, posture change and clothing change. After obtaining the wireless signal, we can reduce the computational overhead by searching on its trajectory. Extensive experiments on WP-ReID dataset demonstrate the effectiveness of our approach. 

\section{Acknowledgments}

This work was supported in part to Dr. Wengang Zhou by NSFC under contract No. 61822208 \& 61632019 and Youth Innovation Promotion Association CAS (No. 2018497), and in part to Dr. Houqiang Li by NSFC under contract No. 61836011.

\bibliographystyle{ACM-Reference-Format}
\bibliography{egbib}

\end{document}